\documentclass[11pt,a4paper]{article}
\usepackage[hyperref]{acl2019}
\usepackage{times}
\usepackage{latexsym}
\usepackage{graphicx}
\usepackage{subfigure}
\usepackage{stfloats}
\usepackage{url}
\usepackage{comment}

\aclfinalcopy % Uncomment this line for the final submission
\def\aclpaperid{2414} %  Enter the acl Paper ID here

\title{Determining Relative Argument Specificity and Stance\\ for Complex Argumentative Structures}

\author{Esin Durmus \\
  Cornell University \\
  \texttt{ed459@cornell.edu} \\\And
  Faisal Ladhak \\
  Amazon\\
  \texttt{faisall@amazon.com} \\\And
  Claire Cardie \\
  Cornell University \\
  \texttt{cardie@cs.cornell.edu}
  }

\date{}

\begin{document}
\maketitle
\begin{abstract}
Systems for automatic argument generation and debate require the ability to
%present a diverse and comprehensive set of supporting and opposing arguments given a controversial topic.
%
%Such systems must, as a matter of course, be able to
(1) determine the stance of any claims employed in the argument and (2) assess the specificity of each claim relative to the argument context.
%ctc: "specificity" in the sense of determining its position relative to 
% other claims along an argument path
%remaining aware of the context at different levels of argumentation. 
Existing work on understanding claim specificity and stance,
however, has been limited to the study of argumentative structures that
are relatively shallow, most often consisting of a single 
%--- automatically generated arguments rarely produce more than a single %supporting/opposing argument for the main claim. 
claim that directly supports or opposes the argument thesis.
%h
In this paper, we tackle these tasks in the context of complex arguments on a diverse set of topics.  
%develop methods to support the generation of diverse and potentially complex arguments on a topic of choice.
In particular, our dataset consists of  manually curated argument trees for $741$ controversial topics covering  95,312 unique claims; lines of argument are generally of depth 2 to 6.  %Claims can either support or oppose the associated parent claim. 
%The root of each tree represents the thesis of the argument, i.e., the issue being discussed; internal nodes represent claims provided in support of, or opposition to, their associated parent claim. 
%Overall, there are $95,312$ unique lines
%of argument in the dataset, most of which range from a depth of 2 to 6 in
%terms of 
%Thus, the dataset encodes a diverse and fairly comprehensive collection of
%arguments that address a broad 
%supporting and opposing points to address each facet of the argument at different depths of the argument tree. 
%
%With then formulate prediction tasks 
%to ultimately support argument generation and debate systems in
%for assessing claim specificity and claim stance. 
%Given any pair of claims along a path in an argument tree, determine their relative depth (specificity); and given claim $A$ at depth $d$ and claim $B$ at depth $> d$ along the same argument path, determine whether $B$ (in)directly \textsc{supports} or \textsc{opposes} $A$ (stance).
We find that as the distance between a pair of claims increases along the argument path, determining the relative specificity of a pair of claims becomes easier and determining their relative stance becomes harder. %Conversely, it is harder to determine their relative stance.
\end{abstract}
% thesis = statement of the topic or question being discusses
% claim/point = statement provided in support of, or opposition to, another claim or the argument thesis
% argument tree 
%   each node is a claim 
%   root node: the thesis (main claim)
%   
%   edges A -> B: two types
%     support: B provided in support of A
%     opposition: B provided in opposition to A
% argument path: sequence of nodes and edges connecting a node to a descendant 
\begin{figure*}[tbhp]
\includegraphics[scale=0.39]{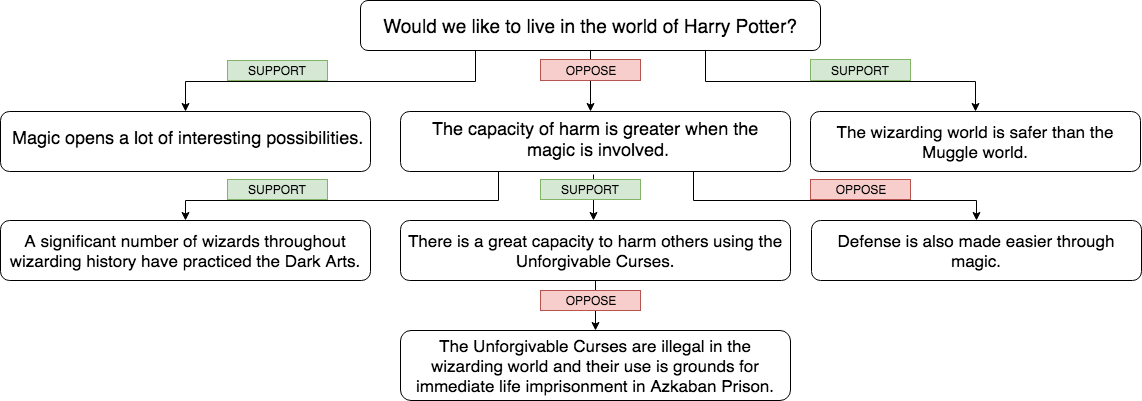}
\caption{Partial tree for the controversial topic ``Would we like to live in the world of Harry Potter?''. Each claim's position towards its parent argument is indicated in the box on the edge between the claim and its parent. The full argument tree for this topic can be found at \href{https://www.kialo.com/is-the-world-of-harry-potter-really-the-place-to-be-2415/2415.0=2415.1}{https://www.kialo.com/is-the-world-of-harry-potter-really-the-place-to-be-2415/2415.0=2415.1}.}
\label{example_data}
\end{figure*}  

\section{Introduction}
The  tasks  of  automatic  argument  generation and debate  require  the  ability  to  present  a diverse  and  comprehensive  set  of  supporting and opposing  arguments  given  a  controversial topic. 
Two critical components of such systems are an ability to determine the \textbf{stance} and the \textbf{specificity} of any claims employed in the
proposed argument. 
Consider, for example, the argument thesis (i.e., the topic) of Figure~\ref{example_data}: (\textsc{Thesis}) 
\textit{Would we like to live in the world of Harry Potter?}  
Construction of an argument in support or in opposition to this thesis
necessarily requires knowing the stance of the claims that comprise it:
the claim \textit{Magic opens a lot of interesting possibilities} 
should be identified as a claim in support of the \textsc{Thesis}, and
\textit{The capacity of harm is greater when magic is 
involved} (\textsc{Harm}), as a claim in opposition. Indeed, previous work has studied this task (e.g., \newcite{E17-1024,Faulkner2014AutomatedCO}).

It is not sufficient, however, to determine claim stance only with respect to the argument thesis.
Debate and argument generation systems, in general, should also be able to determine whether two 
claims that address the same line of reasoning represent the same, or the opposing stance: 
%a given claim supports or opposes another claim along the same line of argument
using \textit{Defense is also made easier through magic} to refute the \textsc{Harm} claim
in Figure~\ref{example_data}, for example, requires recognizing that it represents the opposite stance.

The issue of claim specificity in argumentation has been much less addressed. Existing work, however, suggests that a high
degree of specificity is correlated with argument
quality  and  persuasiveness \cite{P18-1058,W15-4631}. 
In terms of argument quality though,
it is entirely possible  for  the presented claims  to be  coherent and meaningful, yet be too specific within the  given discourse, and therefore be logically irrelevant \cite{dessalles}.  As a concrete example, suppose we wanted to
assert  a  claim  in  support  of  the argument \textsc{Thesis}
of Figure 1.  While \textit{The Unforgivable
Curses are illegal...and their use  is  grounds  for immediate  life  imprisonment} supports the \textsc{Thesis},
it is too specific a claim to introduce at this point in the
argument.  Namely, it doesn't flow naturally without first
introducing the concept of \textit{Unforgivable Curses}.

To date, existing work on understanding claim specificity and stance 
has mostly employed annotated monologic persuasive documents or 
discussion forums and, as a result has been limited to the study
of argumentative structures that
are relatively shallow, most often only consisting of 
%--- automatically generated arguments rarely produce more than a single %supporting/opposing argument for the main claim. 
claims that directly support or oppose the argument thesis \cite{E17-1024,Faulkner2014AutomatedCO}.

To support the generation of diverse and potentially complex arguments on a topic of choice, we present here a dataset of  manually curated argument trees for $741$ controversial topics covering  95,312 unique claims. 
In contrast to existing datasets, ours
consists of argument trees where each root node represents the
argument thesis (main claim) and every other node represents a claim that either supports or opposes its parent. 
%With this structure, we have a diverse set of arguments for each argument thesis as well as for each aspect of the arguments.
Taking advantage of this relatively complex argumentative structure, 
we formulate two prediction tasks to study relative specificity and stance. The main contributions of our study are the following:
\begin{itemize}
    \item We provide a publicly available dataset of argument trees consisting of a diverse set supporting and opposing claims for 
    $741$ controversial topics\footnote{The
dataset will be made publicly available at \href{http://www.cs.cornell.edu/~esindurmus/}{http://www.cs.cornell.edu/~esindurmus/}.}.
    \item We propose two novel settings to study claim specificity and stance in the context of a diverse set of supporting and opposing points.
    \item We control for specific aspects of the argument tree (e.g., depth, stance) in our experiments to understand their effect on claim specificity and stance detection.  
\end{itemize}

\section{Dataset}
We extracted argument trees for $741$ controversial topics from \href{www.kialo.com}{www.kialo.com}\footnote{This covers all controversial topics on the website at the time we collected the data.}.
Kialo is a collaborative platform where users provide supporting and opposing claims for each claim related to a controversial issue. Besides providing the claims themselves, users also help to improve the quality of existing claims by suggesting edits, and rating the quality of claims. This process of collaborative editing helps to create a high quality, diverse set of supporting and opposing points for each controversial topic\footnote{The data is crawled from this website in accordance with the terms and conditions.}.

\begin{figure*}[t]
\centering
\subfigure[Number of controversial topics with the given pre-defined categories. Note that a controversial topic could be related to multiple pre-defined categories.] {\includegraphics[scale=0.32]{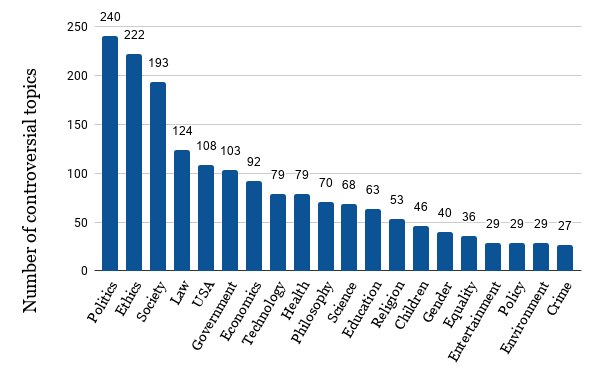} \label{fig:topic_dist}}\quad
\subfigure[Number of claims at given depths. The majority
of the claims lie at the depth 3 or higher.]{\includegraphics[scale=0.30]{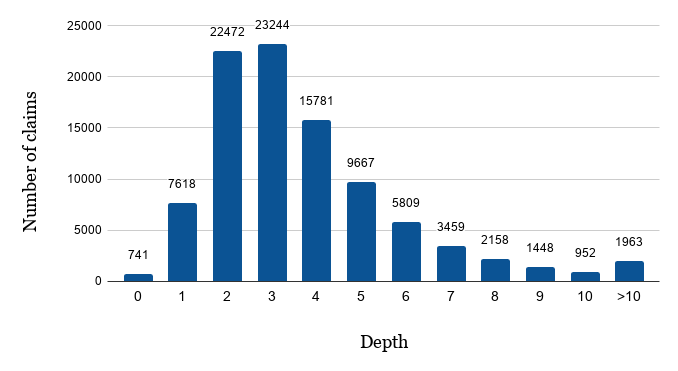} \label{fig:num_node_level}}
\subfigure[Number of trees with given range of total number of claims. For the majority of trees, the argument tree has more than $30$ claims in the tree. Average number of claims per argument tree is 127.] 
{\includegraphics[scale=0.30]{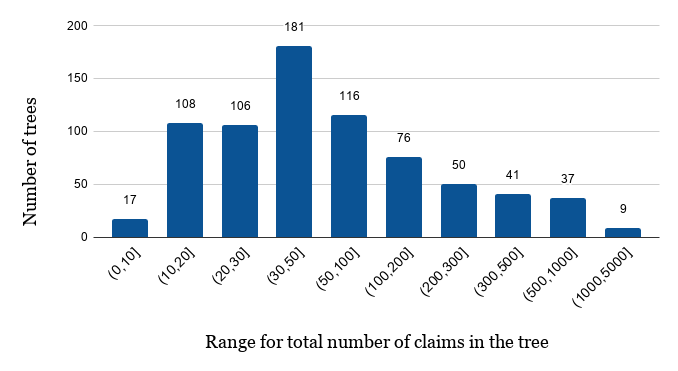} \label{fig:num_node}}\quad
\subfigure[Number of trees with given range of depth. For the majority of trees, the depth of the argument tree is $4$ or higher, and average depth per argument tree is 5.]{\includegraphics[scale=0.30]{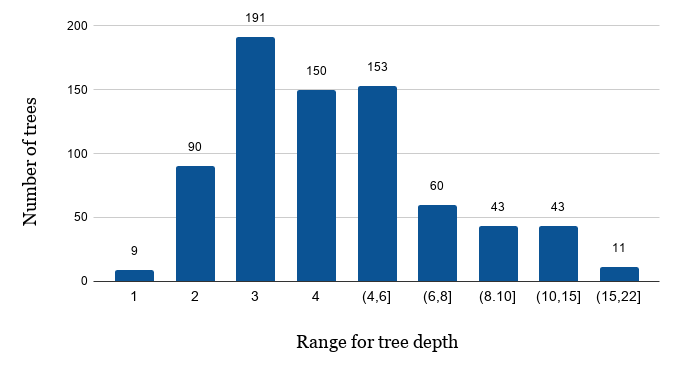} \label{fig:depth}}
\label{fig:data_stat}
\end{figure*}

The dataset includes diverse set of controversial topics. Each controversial topic is represented by a \textbf{thesis} and tagged to be related to pre-defined generic categories such as \textit{Politics},  \textit{Ethics}, \textit{Society} and \textit{Technology}\footnote{Note that a controversial topic can be relevant to multiple pre-defined categories.}. Figure \ref{fig:topic_dist} shows the number of controversial topics with the given pre-defined categories. The controversial topics' theses include: ``A free Press is necessary to democracy.'', ``All drugs should be legalised.'', ``A society with no gender would be better.'', ``Hate speech should be banned'', etc.
\subsection{Structure of the arguments}
The arguments for each controversial topic are represented as trees. The root node of each such tree represents the \textbf{thesis} of the controversial topic. Every other node in the tree represents a \textbf{claim} that either \textbf{supports} or \textbf{opposes} its parent claim. Figure \ref{example_data} shows a partial argument tree for the thesis ``Would we like to live in the world of Harry Potter?''. We see that besides the supporting and opposing claims for the thesis, there are supporting and opposing claims for the claims at different depths. With this structure, we can identify indirect support/oppose relationships even between nodes without parent-child relationships if they are on the same \textbf{argument path}. For example, the claim ``Defense is also made easier through magic'' indirectly supports the thesis, since it is in opposition with its parent ``The capacity of harm is greater when the magic is involved'', which is an opposing claim to the thesis. Another observation is that as we go deeper along an argument path, the claims get more specific, since each claim aims to either support or oppose its parent.
For example, while the claim ``The capacity of harm is greater when the magic is involved'' refers to the general harms that can be caused by magic, one of its child claims ``There is a great capacity to harm others using the Unforgivable Curses'' is more specific as it refers to harm via a particular set of \textit{curses} in \textit{magic}.
\subsection{Data Statistics}
The dataset consists of argument trees for 741 controversial topics comprised of $95,312$ unique claims. The distribution of argument trees with the given range of total claims, and depth is shown in Figures \ref{fig:num_node} and \ref{fig:depth} respectively. We see that for the majority of trees, the depth is $4$ or higher, and the number of claims is greater than $30$. 

Figure \ref{fig:num_node_level} shows the total number of claims at a given depth. We see that only $7,618$ out of $95,312$ claims are directly supporting or opposing the theses of the controversial topics. The majority of the claims lie at the depth $3$ or higher. This shows that the dataset has a rich set of supporting and opposing claims for not only for the theses, but for claims at different depths of the tree.

In total, there are 44,572 claims that are supporting and 50,740 claims that are opposing their parent claims. 90\% of claims consist of $1$ $(61\%)$ to $2$ $(29\%)$ sentences and average number of tokens per claim is $30$.

\begin{figure*}
\centering
{\includegraphics[width=9cm,height=10cm]{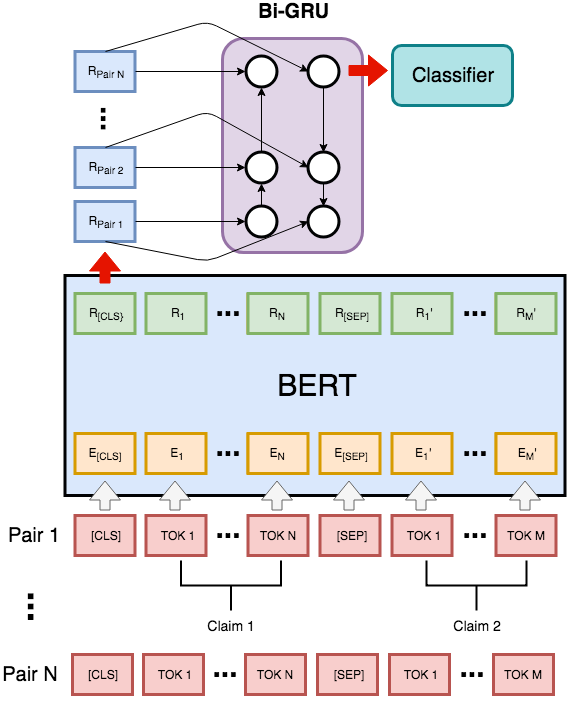}}
\caption{Hierarchical model for stance classification. A pre-trained BERT model is used to encode pairs of claims, which are then fed into a bi-directional GRU, to encode the path. In the figure, E$_i$ represents the input embedding for token TOK$_i$, R$_i$ represents the contextual representation for token TOK$_i$ from the final layer in the BERT model, and R$_{pair\ i}$ is the representation of Pair i.}
\label{fig:bert_model}
\end{figure*}

\section{Claim Specificity}
Determining the relative specificity of arguments is an important step towards being able to generate logically relevant arguments in a given discourse \cite{dessalles}. For a system that disregards the relative specificity of claims, it is entirely possible to generate coherent and meaningful, yet logically irrelevant claims, when the generated claims are either too generic or specific for the given argument discourse. 

In this work, we determine the relative specificity between a pair of claims that are along the same argument path from the thesis to a given leaf claim. We note that specificity always increases along a given path, as each child claim is addressing some aspect of its parent claim, by either supporting or opposing, and therefore by definition has to be more specific. While an increase in depth is correlated with an increase in specificity for claims within a given argument path, this correlation does not necessarily hold for claims across different argument paths within a tree\footnote{We also cannot guarantee that these claims are completely irrelevant and specificity comparison is not applicable. We would need human annotation for these cases to be able to make any claims for the relative specificity.}. One important note is that we use the path information only as a way to reliably generate specificity labels, without requiring human annotations. The task of relative specificity detection itself does not require any path information to be present, nor do we make any assumptions in our models about the availability of path information.

For this task, given a pair of claims, we want the model to determine whether the second claim is more specific than the first claim. We note that unlike in stance prediction, we never provide the path information between a pair of claims, as this would be equivalent to giving the gold label as input to the model, since given the path, the relative specificity is deterministic.

\subsection{Results and Analysis} \label{spec_results}
\textbf{Baseline.} We experiment with feature-based Logistic Regression (LR) model that incorporates all the features that are shown to be effective in determining sentence specificity \cite{generic_spec}. For example, this feature list includes polarity of the claims \cite{Wilson:2005:RCP:1220575.1220619}, number of personal pronouns in the claims, and length of the claims since \cite{generic_spec} shows that generic sentences have stronger polarity, less number of personal pronouns and are shorter in length. While \newcite{ko2019domain} has also looked at the task of specificity prediction, we cannot directly apply their models to our data, since their annotation scheme requires each sentence to be labelled as general or specific, whereas we argue that specificity is relative. 

\textbf{Fine-tuned BERT.} We compare our baselines with a fine-tuned BERT model \cite{devlin2018bert}. BERT is a pre-trained deep bidirectional transformer model that can encode sentences into dense vector representations. It is trained on large un-annotated corpora such as Wikipedia and the BooksCorpus \cite{DBLP:conf/iccv/ZhuKZSUTF15} using two different learning objectives, namely masked language model and next sentence prediction. These learning objectives together allow the model to learn representations that can be easily fine-tuned to achieve state-of-the-art performance for a wide range of natural language processing tasks.

For relative specificity detection, we feed the pair of claims as a single sequence with the special [SEP] token between the claims, and a [CLS] token at the beginning of the sequence, as shown in Figure \ref{fig:bert_model}, into a pre-trained BERT model\footnote{Specifically, we use the BERT-Base (Uncased) model, which contains 12 layers of bidirectional transformers, with a hidden size of 768 units and 12 attention heads (for a total of 110M parameters).}. In addition, we indicate each token in the first claim (as well as the [CLS] and [SEP] tokens) as belonging to sentence A, and each token in the second claim as belonging to sentence B, which is used by the BERT model to add the appropriate learned sentence embedding to each token. Note that this approach of packing a pair of claims into a single sequence is consistent with the input representation from \cite{devlin2018bert}, for tasks where the input is a pair of sequences. We then take the output of the [CLS] token from the final layer of the BERT model, and feed it into a classification layer. We fine-tune\footnote{For all fine-tuning experiments with BERT, we used a learning rate of 2e\textsuperscript{-5}. We ran the fine-tuning jobs for a maximum of 5 epochs, and used the validation performance for early stopping.} this architecture for relative specificity detection.
\begin{table}[]
    \centering
    \begin{tabular}{|l|c|c|c|}
    \hline
        & Train & Development & Test \\
        \hline
      Specificity  &  196,474 & 77,599 & 79,394\\
      \hline
      Stance  & 159,726  & 60,891 &  65,732\\
       \hline
    \end{tabular}
    \caption{Number of examples (claim pairs) in each split for claim specificity and claim stance tasks.}
    \label{table:splits}
\end{table}

We split our data into train, development and test sets, by topic, which ensures that all nodes from the same tree are confined to a single split. We split the data in this way in order to encourage our models to learn more domain independent features, that are applicable across the diverse set of controversial topics. Number of examples in each split for each task is shown in Table \ref{table:splits}.

\begin{table*}
\begin{center}
\begin{tabular}{ |l|c|c|c|c| } 
\hline
Model & All pairs & Distance one & Same stance \\
\hline
\hline
Majority & 50.14 & 50.25 & 49.97 \\
\hline
Length & 74.94  & 64.67 & 69.62 \\
\hline
Bag of Words (BOW) LR & 77.10 & 66.01 & 70.43 \\
\hline
Feature-based LR & 78.18 & 67.06 & 72.03 \\ 
\hline 
BOW + Feature based LR & 79.12 & 67.54 & 73.14  \\
\hline
Fine-tuned BERT & \textbf{84.91} & \textbf{74.51} & \textbf{80.23} \\ 
\hline
\end{tabular}
\end{center}
\caption{Accuracy numbers for argument specificity, across the different settings.}
\label{table:specificity_results}
\end{table*}

\begin{table*}
\begin{center}
\begin{tabular}{ |l|c|c|c|c|c| } 
\hline
Model & d=1 & d=2 & d=3 & d=4 & d=5 \\
\hline
\hline
Length & 64.67 & 76.40 & 80.22 & 80.40 & 79.69 \\
\hline
BOW + Feature based LR & 67.54 & 79.98 & 84.46 & 85.14 & 85.66 \\
\hline
Fine-tuned BERT & \textbf{74.51} & \textbf{85.57}  & \textbf{89.30} & \textbf{90.57} & \textbf{91.62} \\ 
\hline
\end{tabular}
\end{center}
\caption{Accuracy numbers for argument specificity at distance 2-5.}
\label{table:specificity_distance}
\end{table*}

Table \ref{table:specificity_results} compares the performance of the different models for relative specificity, across three different settings. In the first setting, we evaluate the models across all claim pairs that occur in the same argument path in a given tree. We then control for the distance between the pair, in the second setting, by evaluating only across pairs of nodes that are distance 1 from each other, i.e. have a parent-child relationship. Finally, we control for the stance, in the third setting, and evaluate across pairs of claims that have the same stance relative to their parent. 
 
\textbf{Analysis.} Consistent with previous work \cite{Li2015FastAA}, we find that length is highly predictive of specificity and more specific claims are longer than more generic claims. Across all settings, the fine-tuned BERT model achieves the best performance. As expected, the performance degrades, for all models, as we control for distance and stance, since the claims get more similar in language, for both cases. 
 
 Table \ref{table:generic_specific_words} shows the top weighted words by BOW model for each class. We find that connectives (such as also, but, because, when) are associated more with arguments with higher specificity as they are mostly used to add more specific information to the claims as also found by \newcite{W17-5006-Lugini}, whereas concept words (such as society, world, gender) have higher association with more generic arguments since these words represents the concepts of the controversial topics that people argue about. 
\begin{table}
\begin{center}
\begin{tabular}{ |c|c| } 
\hline
More generic & More specific\\
\hline
\hline
 should & also\\ 
\hline
society & but \\ 
\hline
gender &  only \\ 
\hline
world &  because \\
\hline
humans & at\\ 
\hline
rights & when\\ 
\hline
would & even \\ 
\hline
government & that\\ 
\hline
\end{tabular}
\end{center}
\caption{Words associated with more generic and specific arguments.}
\label{table:generic_specific_words}
\end{table}

We further evaluate our models for the claim pairs with distance values $2$ to $5$ as shown in Table \ref{table:specificity_distance}. We find that BERT model is consistently the best performing model for all distance pairs. As we increase the distance, the models achieve higher prediction performance despite having less training examples for higher distance values. 

\section{Claim Stance Detection}
It is not sufficient for debate and argument generation systems to determine the claim stance only with respect to the argument thesis; it is also necessary to determine the stance between any pair of claims that address the same line of reasoning.
An argument generation system, for example, may need to generate arguments that oppose some of the opponent's previous claims while supporting some of its own previous claims during the debate which would require to determine the stance between any candidate claims and the claims in the previous argument discourse.

In this work, given a claim $A$ at
depth $d$ and claim $B$ at depth $> d$ along the same argument path, we
determine whether $B$ (in)directly \textsc{supports} or \textsc{opposes} $A$ (stance). If $A$ and $B$ do not have parent-child relationship, we determine whether $B$ indirectly \textsc{supports} or \textsc{opposes} $A$ by considering support/oppose relationship of each parent-child claims between $A$ and $B$. Following the example shown in Figure \ref{example_data}, the claim ``The capacity of harm is greater when the magic is involved'' is directly supported by the claim ``There is a great capacity to harm others using the Unforgivable Curses'', with a 
\textbf{direct} parent-child relationship. However, the argument ``The Unforgivable Curses are illegal in the wizarding world and their use is grounds for immediate life imprisonment in Azkaban Prison'' is \textbf{indirectly} \textbf{opposing} the same claim, by rebutting it's parent, which presents a supporting point for the claim.
\begin{table*}
\begin{center}
\begin{tabular}{ |l|c|c|c| } 
\hline
Model  &  Distance one  & All pairs \\
\hline
\hline
Majority   & 44.63  & 49.48\\ 
\hline
Feature-based LR    & 63.02  &  55.10 \\ 
\hline
Feature-based LR with path    &  61.27 &  54.70\\ 
\hline
Fine-tuned BERT    &  74.84 &  64.08\\
\hline
Fine-tuned BERT with path (simple)  & 76.77 &  66.22\\ 
\hline
Fine-tuned BERT with path (hierarchical)  & \textbf{77.46} &  \textbf{68.55}\\ 
\hline
\end{tabular}
\end{center}
\caption{Accuracy numbers for argument stance detection, across the different settings.}
\label{table:stance_results}
\end{table*}
\subsection{Results and Analysis}
We experiment with a feature-based Logistic Regression model and a fine-tuned BERT model \cite{devlin2018bert} using the same strategy to split the data into train, development and test sets as in Section \ref{spec_results}. 

\textbf{Baseline.} Our feature-based model employs features shown to be effective in stance detection tasks \cite{S16-1003} such as bag of words, word match, sentiment match, document embedding similarity, and MPQA subjectivity features \cite{Wilson:2005:RCP:1220575.1220619}\footnote{For Featured-based LR with path, we concatenate the all claims along an argument path, and extract features from this concatenated sequence.}. We cannot evaluate the model from \newcite{C18-1203} as a baseline, as that requires additional annotations for argument phrases for the given topics. Similarly, we cannot evaluate the model from \newcite{E17-1024} as a baseline, since it would require additional annotations for target phrases in each claim, polarity towards the target phrases, and consistent/contrastive labels between the target phrases of two claims.

\textbf{Fine-tuned BERT.} We feed a pair of claims into a pre-trained BERT model, in the same manner as detailed above for relative specificity detection, and take the output of the [CLS] token from final layer and feed it into a classifier. We fine-tune this model for relative stance detection. 

\textbf{Fine-tuned BERT with path (simple).} In this model, we incorporate path information in a very na\"{i}ve manner. For a given pair of claims A and B, where A is a predecessor of B, we concatenate the path of claims starting from B up to A with each claim separated by the special [SEP] token. We indicate each token from claim B as belonging to sentence A, and the tokens from all other claims in the path, including claim A, are indicated as belonging to sentence B. We note that this way of processing the input is similar to how \cite{devlin2018bert} processed their input for the QA task. Similar to the previous model, we feed the output of the [CLS] token from the final layer into a classifier. We then fine-tune this model for relative stance classification. 

\textbf{Fine-tuned BERT with path (hierarchical).} We hypothesize that the task of determining relative stance becomes easier, if we can follow along the argument path and determine the relative stance between parent-child claims. We incorporate this inductive bias into the model by constructing a hierarchical architecture for relative stance classification, as shown in Figure \ref{fig:bert_model}. First, we feed each parent-child pair along an argument path as a single sequence into the BERT encoder, separated by the [SEP] token, and take the representation of the [CLS] token from final layer of the BERT model, as the pair representations. We then feed the sequence of pair representations into a bidirectional Gated Recurrent Unit (GRU) \cite{cho-al-emnlp14}, to get the path representation. In our experiments, we used a single bidirectional GRU layer with 128 units. The output of the last token from the forward GRU, and the output of the first token from the backward GRU are concatenated together to get the final path representation. We then feed this into a classifier to predict relative stance. We fine-tune this architecture for relative stance classification.

Table \ref{table:stance_results} compares the performance of the different models for argument stance detection, across two different settings. In the first setting, we evaluate the models only across pairs of claims that are distance 1 from each other, i.e. in a parent-child relationship. In the second setting, we evaluate the model across all pairs that occur in the same argument path in a given tree \textbf{with} and \textbf{without} incorporating the claims along the argument path between these pair of claims. 

\textbf{Analysis.} We find that the fine-tuned BERT models perform much better than the feature based models and baselines, across both the settings. Also, as we hypothesized, having the argument path information is useful for determining relative stance between claims that do not have a parent-child relationship, as the BERT models with path information consistently perform better in the second setting, with the hierarchical BERT model being the best. In our dataset, an argument path from the tree is the best approximation that we have for an argumentative discourse, and as such our results suggest that considering discourse level context is useful in determining relative stance between two claims. However, as shown by our results, our models can still be employed when there is limited or no discourse information.
\begin{table*}[]
    \centering
    \begin{tabular}{|l|c|c|c|c|}
    \hline
    & d=1 & d=2 & d=3 & d=4 \\
    \hline
    Number of examples & 21,451  &
    19,940 & 14,947 & 9,394\\
    \hline
    \hline
    Fine-tuned BERT & 74.84 & 60.69 & 58.34 & 55.88 \\
    \hline
    Fine-tuned BERT with path (simple) & 76.77 & 65.10 & 59.12 & 55.80 \\
     \hline
    Fine-tuned BERT with path (hierarchical) & \textbf{77.46} & \textbf{67.74} & \textbf{62.51} & \textbf{59.51} \\ 
    \hline
    \end{tabular}
    \caption{Accuracy for relative stance at distance 1-4.}
    \label{table:stance_distance_results}
\end{table*}

The performance degrades significantly\footnote{We measure the significance performing t-test.} in the second setting, where we include claim pairs with all the distances, implying that it is easier to determine the stance relative to the parent, than claims that are further on the same path. 

We do a more fine grained analysis of the performance of the fine-tuned BERT models, at different distances, which we present in Table \ref{table:stance_distance_results}. As expected, performance degrades for all models as the distance between the pair of claims increases. We find that at distance d=4 Fine-tuned BERT model that incorporates path information in a simple manner performs similarly to the model without path information. The hierarchical model, however, performs significantly better, which further justifies our choice to treat the argument path context as a hierarchical rather than a flat representation.

\section{Related Work}
\textbf{Argumentation Generation.} Previous work in argument generation has focused on generating summaries of opinionated text \cite{N16-1007}, rebuttals for a given argument \cite{W00-1406}, paraphrases from predicate/argument structure \cite{W03-1601}, generation via sentence retrieval \cite{P15-4019} and developing argumentative dialogue agents \cite{W18-5215,DBLP:journals/corr/abs-1709-03167}. The work on developing argumentative dialogue agents, in particular, has employed mostly social media data such as IAC \cite{Walker2012ACF} to design retrieval-based or generative models to make argumentative responses to the users. These models, however, employ very limited context in generating the claims, and there is no notion of generating a claim with a particular stance or the appropriate level of specificity within the context. Furthermore, these models are trained on social media conversations, which can be noisy, and as noted by \newcite{DBLP:journals/corr/abs-1709-03167}, many sentences either do not express an argument or cannot be understood out of context. In contrast, our dataset explicitly provides the sequence of claims in an argument path that leads to any particular claim, which can enable an argument generation system to generate relevant claims, with a particular stance and at the right level of specificity. Recent work by \newcite{P18-1021} studies the task of generating claims of a different stance for a given statement, however their context is limited to the given statement and they do not take specificity into account.

\textbf{Stance Detection.} 
Previous work on claim stance detection has studied the important linguistic features to determine the stance of a claim relative to a thesis/main claim \cite{P09-1026,W10-0214,N12-1072,WALKER2012719,I13-1191,W14-2715,W06-1639,D10-1102,P11-1151,Kwon:2007:ICS:1248460.1248473,Faulkner2014AutomatedCO,E17-1024}. Some of these studies have shown that simple linear classifiers with uni-gram and n-gram features are effective for this task \cite{W10-0214,I13-1191,S16-1003}. However, in our setting, since we try to predict the stance between all pairs of claims on an argument path, rather than simply claims that are directed towards the thesis or the parent claim, we find that the models with a hierarchical representation of the argument path, i.e. the context, significantly outperform these baselines.

\textbf{Argument Structure and Quality.} 
%Recent work in NLP on argumentation has focused on determining the structure and the quality of the arguments. 
There has been tremendous amount of work in computational argumentation mining focusing on determining argumentative components \cite{mochales2011argumentation,stab-gurevych:2014:Coling,nguyen-litman:2015:ARG-MINING} and argument structure in text \cite{palau2009argumentation,biran2011identifying,feng2011classifying,lippi2015contextindependent,park-cardie:2014:W14-21,peldszus-stede:2015:EMNLP,P17-1091,rosenthal2015couldn}, and understanding the argument quality dimensions \cite{E17-1017,P18-1058} and the characteristics of persuasive arguments \cite{kelman1961processes,burgoon1975toward,
chaiken1987heuristic,tykocinskl1994message,
doi:10.1177/0741088396013003001,N18-1094, dillard2002persuasion, Cialdini.2007, durik2008effects,
tan2014effect, Marquart2016,DBLP:conf/www/DurmusC19}. Existing work on claim specificity and stance detection has mostly employed datasets extracted from monologic documents that include more shallow support/oppose structures \cite{E17-1024,Faulkner2014AutomatedCO}. Although there has been some work on constructing argument structure datasets using news sources \cite{reed-etal-2008-language}, microtexts \cite{peldszus-2014-towards} and user comments \cite{PARK18.679}, these structures tend to be shallower and include fewer opposing claims since they employ existing monologic texts that are relatively short. In contrast, the dataset we provide is constructed with the goal of providing supporting and opposing claims for each of the claim presented in an argument tree. Therefore, these argument tree structures are deeper and have more balanced number of supporting and opposing claims. 

\section{Conclusion}
We present a new dataset of manually curated argument trees, which can open interesting avenues of research in argumentation. We use this dataset to study methods for determining claim stance and relative claim specificity for complex argumentative structures. We find that it is easier to predict stance for claims that have a parent-child relationship, where as relative specificity is more difficult to predict in the same case. For future work, it may be interesting to understand which other models would be effective in claim specificity and stance detection tasks. Besides, developing techniques to incorporate the claim stance and specificity detection models in argument generation to generate more coherent and consistent arguments is another interesting research direction to be explored.

\section*{Acknowledgments} 
This work was supported in part by NSF grants IIS-1815455 and SES-1741441.  The views and conclusions contained herein are those of the authors and should not be interpreted as necessarily representing the official policies or endorsements, either expressed or implied, of NSF or the U.S.\ Government.
\bibliography{acl2019}
\bibliographystyle{acl_natbib}
\appendix
\end{document}